\documentclass{article}

\usepackage{PRIMEarxiv}

\usepackage[utf8]{inputenc} 
\usepackage[T1]{fontenc}    

\usepackage{nicefrac}       
\usepackage{microtype}      
\usepackage{lipsum}
\usepackage{fancyhdr}       
\usepackage{amsmath,amssymb,amsfonts}
\usepackage{textcomp}
\usepackage{xcolor}
\usepackage{tipa}
\usepackage{rotating}
\usepackage{booktabs}
\usepackage{subcaption}
\usepackage{lineno}
\usepackage{soul}
\usepackage{multirow}
\usepackage{hyperref}
\usepackage{tablefootnote}
\usepackage{xurl}
\usepackage{algorithm}
\usepackage{algpseudocode}

\usepackage{graphicx}
\graphicspath{ {./images/} }

\title{FM-G-CAM: A Holistic Approach for Explainable AI in Computer Vision
}

\author{
  Ravidu Suien Rammuni Silva$^1$, Jordan J. Bird$^2$ \\
  Department of Computer Science \\
  Nottingham Trent University \\
  Nottingham, Nottinghamshire, United Kingdom\\
  \texttt{$^1$ravidu.rammunisilva2022@my.ntu.ac.uk, $^2$jordan.bird@ntu.ac.uk} \\
}

\begin{document}
\maketitle

\begin{abstract}
Explainability is a vital aspect of modern AI for real-world impact and usability. The main objective of this paper is to emphasise the need to understand the predictions of Computer Vision models, specifically Convolutional Neural Network (CNN) models. Existing methods for explaining CNN predictions are largely based on Gradient-weighted Class Activation Maps (Grad-CAM) and focus solely on a single target class; this assumption about the target class selection neglects a large portion of the predictor CNN's prediction process. In this paper, we present an exhaustive methodology, called Fused Multi-class Gradient-weighted Class Activation Map (FM-G-CAM), that considers multiple top-predicted classes and provides a holistic explanation of the predictor CNN's rationale. We also provide a detailed mathematical and algorithmic description of our method. Furthermore, alongside a concise comparison of existing methods, we compare FM-G-CAM with Grad-CAM, quantitatively and qualitatively highlighting its benefits through real-world practical use cases. Finally, we present an open-source Python library with an FM-G-CAM implementation to conveniently generate saliency maps for CNN-based model predictions.
\end{abstract}



\section{Introduction}
\label{sec:introduction}
Explainability is an aspect of Artificial Intelligence (AI) that is gaining prominence today. Explainable AI (XAI) enables humans to understand model predictions more comprehensively than \textit{black box} approaches, which is paramount for the use of such technologies in the real world \cite{arrieta2020explainable, minh2022explainable, loh2022application}. Among many other fields, computer vision often benefits from XAI, in part due to its visual nature \cite{nafisah2022tuberculosis, sudar2022alzheimer, minh2022explainable}. In computer vision, both Convolutional Neural Networks (CNNs) and Transformer-based models are widely used for image classification \cite{alzubaidi2021review,han2022survey, khan2022transformers}. Notably, several transformer-based approaches use convolutional layers.

Despite the strengths of CNNs in the field of computer vision, they are often considered a black-box approach \cite{Guidotti2018camsurvey}, meaning their predictions are difficult to analyse and interpret. In critical situations where machine learning models are used in the real world, the process that leads to a certain decision may be just as important as the model's own ability. The Gradient Class Activation Mapping (Grad-CAM) approach \cite{selvaraju2017grad} was proposed to visually display the convolutional processes in an easy-to-understand manner. The Grad-CAM algorithm generates a saliency map that highlights the areas of a given image that are most important for a classification prediction.

However, relying on a single predicted class to generate these visual explanations introduces vulnerabilities to human-AI teaming and confirmation bias. When a user is presented with a saliency map that shows only the model's top-1 prediction, they are artificially blinded to the underlying uncertainty and to the presence of secondary yet highly relevant visual features. This is particularly problematic in light of recent global directives, such as the European Union Artificial Intelligence Act and the FDA guidelines for Software as a Medical Device (SaMD), which increasingly demand that AI systems provide holistic, transparent, and clinically faithful interpretations of their outputs \cite{EUAIAct2024,amann2020explainability}. A singular viewpoint fails to capture the multi-faceted nature of complex images containing multiple objects or co-occurring pathologies, sometimes resulting in visually compelling but misleading explanations \cite{adebayo2018sanity}.

The Grad-CAM saliency map is based on the activations that the CNN model produces when making a prediction and on their gradients. Building on this foundation, more advanced studies have introduced improved techniques, such as using the second derivative of the resulting activations with respect to the final prediction \cite{chattopadhay2018grad}. However, all of these studies consider only a single class when producing saliency maps. This is commonly the top predicted class, i.e., the one the model considers the most likely label. We argue that, with the exception of binary classification problems, this does not represent the complete rationale of the model for making a given prediction; the class with the highest probability is shown regardless of the context of other prediction values. This is mainly because the saliency map is generated for a class that may or may not even be the desired one. The saliency map is thus highly dependent on the model output with the highest probability. The difference in the top-1 and top-5 accuracy rates of models \cite{alzubaidi2021review, gupta2022deep} trained on ImageNet \cite{deng2009imagenet} argues that even the most accurate models are not always able to predict the correct class as the top prediction. In this study, we present FM-G-CAM, an approach to generate saliency maps that consider multiple predicted classes, providing a holistic visual explanation of CNN predictions.

\begin{figure}[h]
    \centering
    \includegraphics[width=0.7\textwidth]{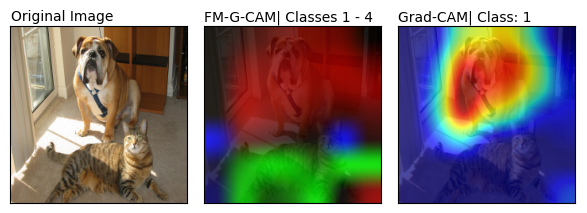}
    \caption{FM-G-CAM for general image classification tasks against Grad-CAM}
    \label{fig:general-comparison-0}
\end{figure}

In this paper, we propose a novel approach to producing a saliency map that offers a unified way to visualise multiple-class predictions; an example of the proposed approach is shown in Figure \ref{fig:general-comparison-0}. Our approach can be configured to consider any number of top classes when generating the saliency map. Our paper begins with a comparison of existing methods for explaining CNN predictions. We then explain our new approach, FM-G-CAM, and the theory behind it, and implement it as a Python package for public download. Following the theory, we present a list of practical use cases for which our approach could be effectively used, compared with the existing methods. Finally, the paper concludes with an in-depth discussion of how FM-G-CAM can be widely used to explain CNN predictions, its drawbacks, and potential avenues to further improve it. The main scientific contributions of our work are as follows:

\begin{itemize}
    \item A concise review of existing saliency map generation techniques. The review discusses similarities, differences, and drawbacks, highlighting the gaps that remain. We explain the three main techniques for explaining CNN predictions and further investigate activation-based saliency map generation methods.
    \item A holistic approach to explaining CNN predictions, with practical use cases. To address one of the main drawbacks highlighted in the review of existing work, a novel technique, Fused Multi-class Gradient-weighted Activation Map (FM-G-CAM), is introduced. FM-G-CAM can consider multiple predictor classes when generating saliency maps, thereby providing a holistic explanation of CNN-based model predictions. We also provide an exhaustive mathematical and algorithmic explanation of FM-G-CAM, discussing in depth the main concepts behind the novel technique. 
    \item A quantitative evaluation based on a customised test plan that repurposes existing saliency map evaluation techniques. The tests use a pre-trained model and comparatively evaluate FM-G-CAM against Grad-CAM across more than 80,000 image samples.
    \item Practical usage examples similar to Grad-CAM, highlighting FM-G-CAM's benefits. The examples cover general use cases and a potential use case for Medical AI in Chest X-ray diagnosis. We used open-source, state-of-the-art models and datasets for the prediction tasks.
    \item A Python package for the convenient generation of saliency maps for CNN predictions. We released this PyTorch-compatible library to the Python Package Index (PyPI), allowing the research community to use FM-G-CAM freely and conveniently via a Python library. The library supports different configurations of FM-G-CAM as well as the original Grad-CAM.
\end{itemize}

\section{Existing Work}
\label{sec:litreview}
In this section, we critically evaluate the existing techniques for generating saliency maps for CNN-based predictions. We begin by discussing different methods for explaining the convolutions that occur in a CNN-based model. We then summarise virtually all published CAM-based techniques, allowing the reader to clearly see the differences among them. Finally, we conclude the review of existing work by discussing the characteristics that make up a good saliency map.

\subsection{Explaining Convolutions}
Convolutional Neural Networks were long considered a black box \cite{Guidotti2018surveryblackbox} mainly because their predictions were difficult to explain logically. Currently, several methods attempt to visually explain the inner workings of CNN-based models, including CNN-based transformer models \cite{SOBAHI20221066, Chefer_2021_CVPR, JUNGIEWICZ2023120234}. These visualisation methods predominantly use model properties such as filter weights \cite{NIPS2012_c399862d, WANG2021203696}, hidden neurons \cite{Rauber2017Visualization, Mahendran2016, Bau2017}, and inference values, which are discussed in the next section.

\subsubsection{Inference values}
Inference values of a CNN represent the inner workings of the model. These values include neuron activations, calculated gradients, the main input, and the main output. However, it is not straightforward to visualise these values in a meaningful way. To date, three main techniques have been used to meaningfully convert these quantitative values into a qualitative saliency map. Table \ref{tbl:cnn-vis-types} presents existing studies, classified by their techniques and drawbacks.
\begin{table}[]
\centering
\caption{Main techniques of CNN prediction visualisation}
\label{tbl:cnn-vis-types}
\resizebox{\textwidth}{!}{%
\begin{tabular}{p{2.8cm}p{12cm}}

\hline
\textbf{Technique} &
  \textbf{Studies} \\ \hline
\textbf{Perturbation} &
Blinded grey patches were used by Zeiler et al. \cite{Zeiler2014} to identify the areas of the image that affect the prediction results. The variation in the targeted class score is recorded and then used to generate a saliency map visualising the areas that are important for the prediction. Later, this technique was further advanced, replacing the grey square with different pixel-value alteration techniques such as masking \cite{zintgraf2017visualizing}.

Although this technique is relatively straightforward, it is very inefficient because each altered patch must be evaluated by inferring the patched image through the model. The technique becomes even more inefficient when there is more than one patched or altered area in the image. \\ 

\textbf{Propagation} &

This technique requires only that the image be propagated once forward and once backward \cite{simonyan2014deep}. The bare gradients calculated during inference are used to generate the saliency map. This technique was much faster than the perturbation-based techniques. However, the generated saliency maps were occasionally very noisy. Later, more studies have been published that address this noise problem by adding slight modifications to the backpropagation algorithm \cite{springenberg2015striving, bach2015pixel, sundararajan2017axiomatic, montavon2017explaining}.

The main drawback of this technique is the noise present in the saliency maps. \\ 
  
\textbf{Activation} &
Saliency maps generated using model activations during inference are commonly called Class Activation Maps (CAMs). This was first introduced by Zhou et al. \cite{zhou2016learning}, who used activations from the feature maps at the penultimate layer. This was done by identifying the relevance of these activations to the final fully connected output for the target class. However, this was only possible when the penultimate layer was a Global Average Pooling layer, which was the case for most typical CNN architectures at the time. Later, Selvaraju et al. \cite{selvaraju2017grad} presented a way to generate saliency maps for any CNN-based model by computing the CAM of the last convolutional layer and weighting it with the average of its gradients. This is currently the most popular method for generating saliency maps. Grad-CAM was further advanced by also using the second-order derivative in Grad-CAM++ \cite{chattopadhay2018grad}. The idea of Gradient-weighted Activations then led to a series of techniques presented in the section \ref{subsec:cam-summary}.

Lower resolution of the saliency map is one of the main drawbacks of this technique. \\ \hline 
\end{tabular}%
}
\end{table}

\subsection{CAM-based techniques in summary}
Class activations have shown prominent results in explaining CNN predictions, mainly due to the information that CNN activations hold regarding the CNN decision-making process in predictions.
Table \ref{tbl:cam-types} presents a summary of the major CAM-based saliency map generation methods widely used.

\label{subsec:cam-summary}
\begin{table}[H]
\centering
\caption{Common techniques for CNN prediction visualisation.}
\label{tbl:cam-types}
\resizebox{\textwidth}{!}{%
\begin{tabular}{p{3.3cm}p{11.5cm}}

\hline
\textbf{CAM Method} &
\textbf{Brief} \\ 
\hline
  
\textbf{Grad-CAM} \cite{selvaraju2017grad} &
Takes the activations of the last convolutional layer and weights them by the average gradients of the same layer with respect to the final fully connected output for the target class
\\ 

\textbf{XGradCAM} \cite{fu2020axiom} &
It follows a very similar technique to Grad-CAM for weighting activations. However, the calculated gradients were normalised before weighting.
\\ 
  
\textbf{HiResCAM} \cite{draelos2020use}&
No improvement is the resolution of the CAM, as the name suggests. However, instead of averaging the calculated gradients, the authors in this study multiply the activations element-wise by their corresponding gradients.
\\ 

\textbf{Seg-XRes-CAM} \cite{Hasany_2023_CVPR}&
A CAM method inspired by HiResCAM for generating saliency maps for segmentation, with a primary focus on object detection.
\\ 

\textbf{Ablation-CAM} \cite{desaiAblation2020}&
Generates the saliency map in a manner similar to an ablation test. It sets the prediction activations to zero and measures the resulting change in the prediction score. This measured deviation, mapped to the corresponding areas of the original image, is then used to create the final saliency map.
\\ 

\textbf{Score-CAM} \cite{wang2020score}&
Score-CAM removes the dependency on gradients for saliency map generation by using a two-stage approach. Instead of gradients, the importance of activations is determined in the second stage by a perturbation method similar to that of Ablation-CAM.
\\ 

\textbf{Eigen-CAM} \cite{muhammad2020eigen}&
Different from most other CAM methods, Eigen-CAM uses the first principal component of the feature maps produced after a prediction. This technique can also generate a saliency map without a target class.
\\ 

\textbf{Layer-CAM} \cite{Jiang2021layercam}&
Very similar to Grad-CAM's core techniques, but can use an arbitrarily selected convolutional layer of the predictor CNN model. 
\\ 
 
\hline 
\end{tabular}%
}
\end{table}

In addition to the aforementioned methods, Collins et al. \cite{collins2018deep} propose a matrix factorisation-based method that segments the image into regions according to predicted classes.

While CAM-based techniques originated with Convolutional Neural Networks, recent advances have seen the rise of Vision Transformers (ViTs) \cite{han2022survey,khan2022transformers}. Explainability in ViTs relies heavily on attention rollouts and relevance mechanisms \cite{Chefer_2021_CVPR, barekatain2025evaluating}. However, CNNs and hybrid CNN-Transformer architectures remain the backbone of many deployment environments, particularly in medical imaging, where localised feature extraction is critical \cite{sarvamangala2022convolutional, kamnitsas2017efficient}. A persistent limitation across both CNN and Transformer XAI methodologies is `class entanglement': the difficulty of visually separating the reasoning for multiple distinct yet simultaneously predicted classes. Current CAM variants \ref{tbl:cnn-vis-types} implicitly assume a winner-takes-all classification paradigm, leaving a significant gap for unified, multi-class visual explanations.

\subsection{What makes up a good saliency map?}
\label{subsec:lit-good-saliencymap}

Reviewing the existing literature, we identified four main areas that could define the quality of a saliency map; activation weighting technique, colourmap type, resolution, and multiclass support.

The activation weighting technique affects the accuracy of the saliency map at a fundamental level. It matters because it is not only the activations that represent the ``thinking'' process of the CNN, but also the gradients that correspond to them. However, these gradients can be noisy and may inflate unrelated activations during the weighting process \cite{Guidotti2018camsurvey}.

Lui et al. \cite{Liu2018jetcol} present a detailed analysis of different types of quantified colour maps used for data visualisation. Current CAM methods generally use \textit{viridis} or \textit{jet} colour maps when presenting the saliency map. Our opinion is that the `viridis' colourmap is more appropriate than `jet' for saliency maps, since the multiple colours in the `jet' map can make the map more confusing. The colourmap \textit{viridis} is also more suitable, since we do not need to represent narrow intensity ranges in the map.

Generally, the resolution of saliency maps is low and depends on the properties of the predictor model. If the predictor model has larger kernels, the resulting saliency maps will also have higher resolution. However, kernels tend to become smaller as the model deepens. Morbidelli et al. \cite{Morbidelli2020} present a way to use image augmentation techniques to increase the resolution of the saliency map while keeping activation maps smaller. This is useful for classification problems with relatively small visual targets.

Lastly, we argue that the saliency map must consider multiple classes if and when available. Most CAM methods focus only on one targeted class and do not represent the whole context of the CNN model's thinking process. To address this issue, we introduce FM-G-CAM.

\section{Method}
\label{sec:method}

We introduce Fused Multi-class Gradient-weighted Class Activation Map (FM-G-CAM), which investigates the reasoning behind model predictions in a broader way than existing methods, thereby aiming to increase explainability. In this section, we describe the background that led to the development of FM-G-CAM, followed by a comprehensive explanation of the proposed FM-G-CAM algorithm.

According to the relevant literature, there is a knowledge gap in XAI for computer vision predictions when more than one class is possible for an input image. Although techniques such as Deep Feature Factorisation \cite{collins2018deep} address this problem, these approaches focus more on dissecting an area classified as a single class into multiple classes. The proposed FM-G-CAM approach aims to provide a more holistic explanation of a CNN prediction without relying on a single class.

\subsection{Theory}
\label{subsec:theory}
Building on the basic concepts of the Grad-CAM approach \cite{selvaraju2017grad}, FM-G-CAM targets multiple classes rather than a single class, thereby affecting the final prediction results. Grad-CAM focuses on a single output at a time, leaving out important information, especially in multi-class classification. FM-G-CAM attempts to overcome this drawback by producing a multicoloured saliency map that highlights the spatial regions of the image corresponding to the top $N^{*}$ predictions. This can be selected arbitrarily; we selected the top 5 classes ($K = 5$) in this paper. We empirically found that $K = 3$ and $K = 4$ work well with the technique.

Here ($N = \left\{ Prediction\ class\ numbers \right\}$), ($n \in N$) and ($N^{*}$) is defined as follows:

\begin{equation}
    \label{eq:N-set-def}
    N^{*} = \{ n_{k}^{*}:\ \ n_{k}^{*} \in N,\ P\left( n_{1}^{*} \right) > \ldots > P\left( n_{K}^{*} \right)\ and\ k\mathbb{\in N,\ }1 \leq k \leq K\}.
\end{equation}

$P(n)$ denotes the predicted probability of the n\textsuperscript{th} class. For example, $n_{1}^{*}$ denotes the class number of the most probable predicted class. $y^{n}$ denotes the output of the final layer for the n\textsuperscript{th} class:

\begin{equation}
    \label{eq:partial-der-activations}
    \alpha_{c}^{n_{k}^{*}} = \ \frac{1}{I \times J}\sum_{i = 1}^{I}{\sum_{j = 1}^{J}{\ \frac{\partial y^{n_{k}^{*}}}{\partial a_{i,j}^{n_{k}^{*}}}}},\ where\ a^{n_{k}^{*}} \in \ \mathbb{R}^{I \times J}.
\end{equation}

In equation \ref{eq:partial-der-activations}, $\alpha_{c}^{n_{k}^{*}}$ represents the importance matrix for the class $n_{k}^{*}$ and the feature channel $c$, while $a$ represents the convolutional activations. This is then weighted according to their corresponding activations:

\begin{equation}
    \label{eq:averaring-activations}
    R^{n_{k}^{*}} = \frac{1}{C}\sum_{c = 1}^{C}{\alpha_{c}^{n_{k}^{*}}a^{c},}\ where\ R^{n_{k}^{*}} \in \ \mathbb{R}^{I \times J}.
\end{equation}

In equation \ref{eq:averaring-activations}, $R^{n_{k}^{*}}$ denotes the unfiltered saliency map for the class $n_{k}^{*}$, while $N$ denotes the number of classes.

The matrix $S^{n_{k}^{*}}$ denotes the filtered saliency map of size $I \times J$ for the class $n_{k}^{*}$. Each element $S_{i,j}^{n_{k}^{*}}$ is defined as follows:

\begin{equation}
    \label{eq:general-solution-s}
    S_{i,j}^{n_{k}^{*}} = \left\{ \begin{aligned}
    R_{i,j}^{n_{k}^{*}},\ \  & R_{i,j}^{n_{k}^{*}} = max(R_{i,j}^{n_{1}^{*}},\ldots,R_{i,j}^{n_{K}^{*}}) \\
    0,\ \  & R_{i,j}^{n_{k}^{*}} \neq max(R_{i,j}^{n_{1}^{*}},\ldots,R_{i,j}^{n_{K}^{*}})
    \end{aligned} \right.\ ,\ \ where\ S^{n_{k}^{*}} \in \ \mathbb{R}^{I \times J}.
\end{equation}

As shown mathematically in the equation above \ref{eq:general-solution-s}, the filtered saliency map contains only the highest value across the selected $K$ classes for each element in the 2D matrix $S^{n_{k}^{*}}$. All other elements are set to zero.

Then, unlike Grad-CAM, $L_2$-normalisation is applied to a concatenated K-dimensional matrix $S^{'}$:

\begin{equation}
    \label{eq:matrix-concat}
    S^{'} = \ \begin{bmatrix}
    S^{n_{1}^{*}} & \ldots & S^{n_{K}^{*}}
    \end{bmatrix},\ \ where\ S^{'} \in \ \mathbb{R}^{I \times J \times K},
\end{equation}

and passed through the ReLU activation function to obtain the final saliency map matrix $S_{FM-G-CAM}$:

\begin{equation}
    \label{eq:fm-gcam-final-eq}
    S_{FM - G - CAM} = ReLU\left( \frac{S^{'}}{\left\Vert S^{'} \right\Vert_{2}} \right).
\end{equation}

As mentioned above, we set $K$ to 5 in this paper for ease of explanation. The choice of $K$ is discussed in Section \ref{subsec:k-val-selection}.

$L_2$-normalisation helps normalise feature maps across classes, making lower activations more pronounced and visible when displayed. We provide a comprehensive visual demonstration and comparison of Grad-CAM and FM-G-CAM in Section \ref{sec:results}.

\subsection{Algorithm}
\begin{figure}[h]
    \centering
    \includegraphics[width=0.99\textwidth]{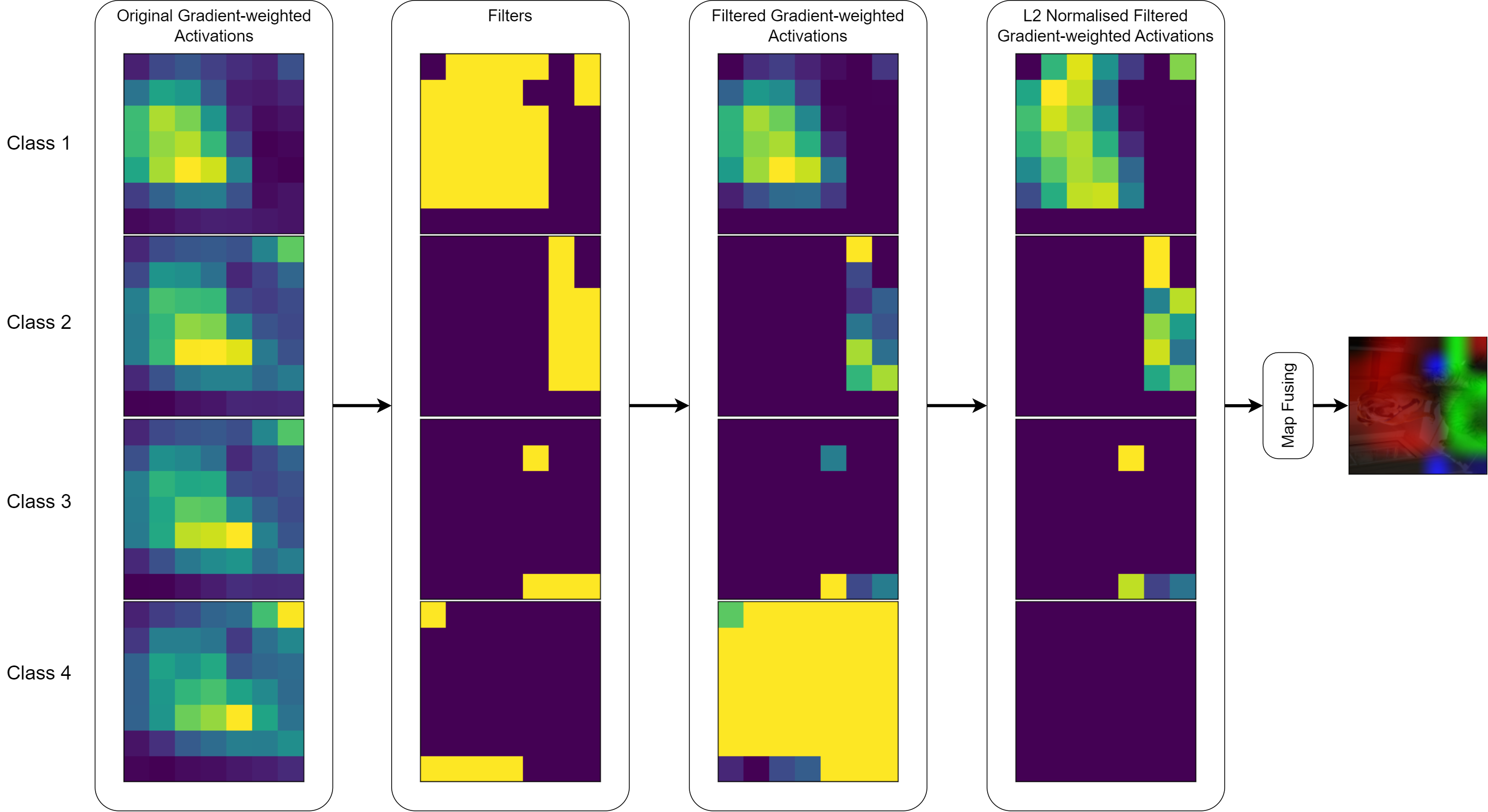}
    \caption{The process of generating FM-G-CAM.}
    \label{fig:fmgcam-process}
\end{figure}

In this section, we present the algorithm that implements the mathematical equations explained in Section \ref{subsec:theory}. This can be used as a guide to implementing this technique from scratch. Figure \ref{fig:fmgcam-process} illustrates the step-by-step generation of FM-G-CAM, from gradient-weighted activation maps to the final FM-G-CAM saliency map.

\begin{algorithm}
\caption{Algorithm for generating FM-G-CAM}
\label{alg:fm-g-cam-generation}
\begin{algorithmic}

\Require $n \geq 1$
\Ensure $c \geq n$
\State $y \gets 1$
\State $C \gets c$
\State $N \gets n$
\State $M = []$

\While{$N \leq C$} \Comment{Calculation of Gradient-weighted Activation for each class}

\State $A_C \gets$ Activations based on class $C$
\State $W_C \gets$ Gradients of activations of class $C$ w.r.t. the selected convolutional layer
\State $M_C \gets A_C*W_C$
\State Append $M_C$ to $M$

\EndWhile
\State Filtering \Comment{see section \ref{subsec:theory}}
\State Normalise $M$ \Comment{$L_2$ Norm}
\end{algorithmic}
\end{algorithm}

\subsection{Choosing an optimal value for (K)}
\label{subsec:k-val-selection}
Unlike traditional saliency methods, FM-G-CAM utilises the $K$ top classes to construct the final fused map. We recommend selecting the top predicted classes. While $K$ can be treated as a static hyperparameter (e.g., $K=3$ or $K=5$), we recommend a dynamic thresholding approach based on the cumulative probability distribution of the model's outputs. Drawing on principles of network calibration and predictive uncertainty \cite{guo2017calibration}, we select $K$ such that the sum of the predicted probabilities exceeds a confidence threshold $\tau$ (e.g., $\sum_{k=1}^{K} P(n_{k}^{*}) \ge 0.95$), allowing the saliency map to adapt dynamically to the prediction's entropy. For images with high certainty and a single focal object, $K$ naturally reduces to 1 or 2; for highly complex or ambiguous scenes, $K$ increases, providing a broader explanation of the model's divided attention.

Based on the above factors, the optimal number of classes for generating saliency maps should be determined. Furthermore, these three factors reveal the complexity of the saliency map required to provide a holistic explanation of the model's predictions.

\subsubsection{Importance of $L_2$ Normalisation}
\begin{figure}[h]
    \centering
    \includegraphics[width=0.75\textwidth]{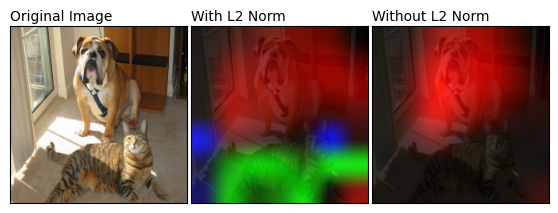}
    \caption{Effect of $L_2$ Norm for saliency map generation in FM-G-CAM.}
    \label{fig:l2-norm-comparison}
\end{figure}

As shown in Figure \ref{fig:l2-norm-comparison}, $L_2$-normalisation across class gradient maps makes FM-G-CAM both sensitive and unbiased, especially when some classes have relatively low-weighted activations, while remaining accurate with respect to the image content. Mathematically, raw gradients from deeper convolutional layers can exhibit severe scale disparities across class predictions. Standard Min-Max scaling or $L_1$-normalisation would cause classes with mathematically dominant but spatially narrow activations to disproportionately overshadow secondary classes. $L_2$-normalisation \ref{eq:fm-gcam-final-eq} acts as a non-linear regulariser, penalising disproportionately extreme activation spikes and smoothing the distribution of the fused maps. This ensures that the qualitative visual map remains faithful to the spatial distribution of features rather than being skewed by raw numerical gradient explosion.

Also, as explained in the algorithm \ref{alg:fm-g-cam-generation}, normalisation is applied only after the filtering process. This allows filtering to be done based on raw gradient-weighted activations, independent of the normalisation process. $L_2$-normalisation also helps to reduce unwanted noise, as shown in Figure \ref{fig:fmgcam-process}. This is because it is applied across the saliency maps by comparing the values against each other.

\subsubsection{Choice of Colour Maps}
Based on the findings presented in Section \ref{subsec:lit-good-saliencymap}, we recommend using a single-colour colour map for each class when K in Section \ref{subsec:k-val-selection} is set to the recommended value of four. This will allow proper visualisation of the CNN decision-making process without overly complicating the saliency map. We compare these in Section \ref{sec:results} with existing methods.






\section{Implementation: XAI Inference Engine}
\label{sec:implementation}

\begin{figure}[]
    \centering
    \includegraphics[width=0.4\textwidth]{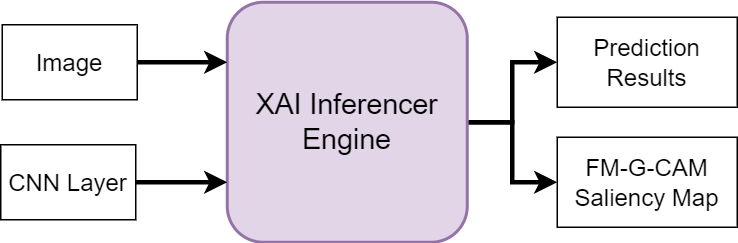}
    \caption{Overview of the XAI Inference Engine.}
    \label{fig:xai-engine-diagram}
\end{figure}

To make FM-G-CAM easy to use, we made a Python library available to the community\footnote{\url{https://pypi.org/project/xai-inference-engine}}. The implementation is based on the algorithm \ref{alg:fm-g-cam-generation}. The code is open source and available on GitHub\footnote{\url{https://github.com/SuienS/xai-inference-engine}}. The library allows users to apply their trained PyTorch CNN-based models. It outputs the prediction results along with the generated FM-G-CAM saliency map explaining the results. Figure \ref{fig:xai-engine-diagram} shows an overview of the XAI Inference Engine.

\begin{figure}[h]
    \centering
    \includegraphics[width=0.99\textwidth]{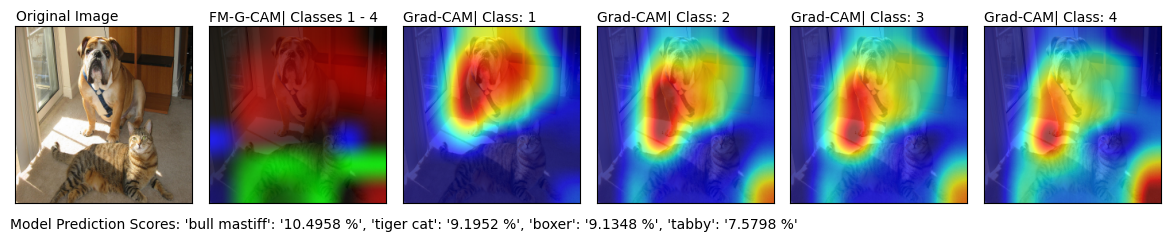}
    \includegraphics[width=0.99\textwidth]{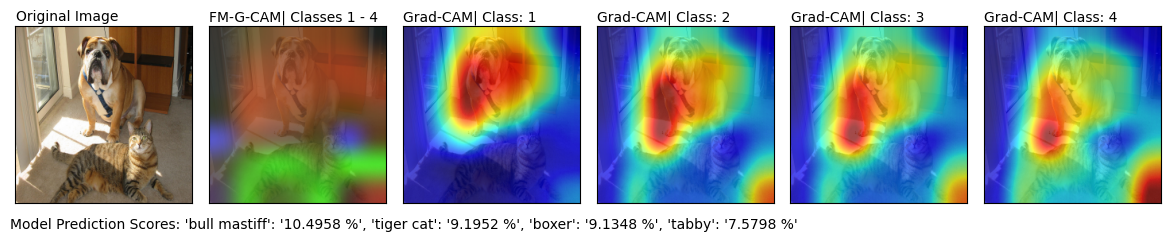}
    \includegraphics[width=0.99\textwidth]{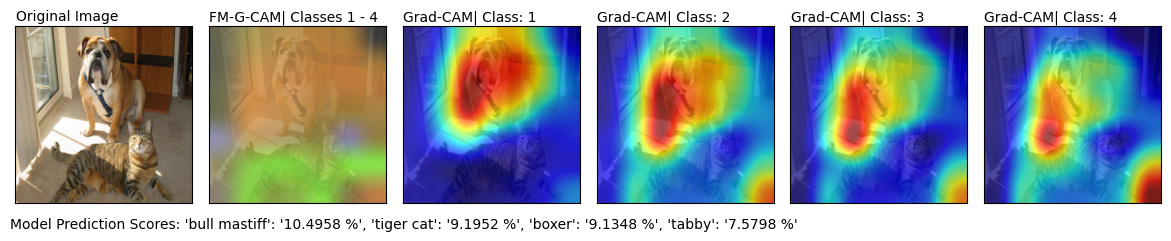}
    \caption{Overview of the XAI Inference Engine. Column 2 shows the output for FM-G-CAM, while columns 3 to 6 show the corresponding saliency map outputs from Grad-CAM.}
    \label{fig:activation-comparison}
\end{figure}

The library also allows the end activation function used in saliency map generation to be changed to ReLU, ELU, or GeLU. Figure \ref{fig:activation-comparison} shows the comparative results of using the different activation functions available. It shows that GeLU and ELU tend to introduce more noise than ReLU. However, they can be useful when we want the saliency maps to be more sensitive. In contrast, Grad-CAM remains identical across all activation functions.

\section{Quantitative Comparison: FM-G-CAM Vs Grad-CAM}
\label{sec:qe-results}
To quantitatively evaluate the effectiveness of FM-G-CAM, we carried out a series of experiments comparing FM-G-CAM with Grad-CAM. Grad-CAM was used in the experiments because we identified it as the theoretical basis even for later techniques such as Grad-CAM++ \cite{chattopadhay2018grad}. 

\subsection{Methodology}
Our tests use four existing metrics to evaluate the reliability of a generated saliency map explaining a given vision-based model prediction. The metrics used in the tests are Deletion Area Under Curve (DAUC) and Insertion Area Under Curve (IAUC) \cite{petsiuk2018rise, petsiuk2021black}, with Insertion Correlation (IC) and Deletion Correlation (DC) \cite{gomez2022metrics}. The DAUC metric is calculated by progressively masking the input image, starting from the most important area for the given class as indicated by the saliency map. IAUC is calculated similarly, but it starts with a blank image and reveals the most important areas sequentially until the entire image is revealed. DC and IC represent the correlation between the target class score and the progressive masking or revealing steps, similar to DAUC and IAUC.

In section \ref{subsec:lit-good-saliencymap}, we discussed the need for saliency maps to be based on multiple classes rather than a single predicted top class. To objectively evaluate whether FM-G-CAM provides a solution to this issue by producing a fusion of multiple class activation maps, we repurposed the aforementioned metrics by calculating them repeatedly across all top-5 predicted classes. With this method, we aimed to assess how these metrics vary with the ranks of the top-5 predicted classes.

To carry out the planned tests, we used the ResNet50V2 \cite{he2016identity} model with ImageNet weights\footnote{https://pytorch.org/vision/main/models/generated/torchvision.models.resnet50.html} and calculated the selected test metrics on the MS-COCO training set \cite{lin2014microsoft}, which contains 80,000+ images. In summary, we used the \textit{knowledge} of ImageNet to classify MS-COCO. MS-COCO was selected for this test because the majority of the images contain more than one ground truth class, which allows the planned test to be carried out meaningfully.
To calculate these metrics, we used an open-source Python library\footnote{https://github.com/TristanGomez44/metrics-saliency-maps} introduced in \cite{gomez2022metrics}. We have also made our test scripts available as open source on GitHub\footnote{https://github.com/SuienS/cam-evaluation}.

\subsection{Results}
\begin{figure}[h]
    \centering
    \begin{subfigure}[b]{0.45\linewidth}
    \includegraphics[width=\linewidth]{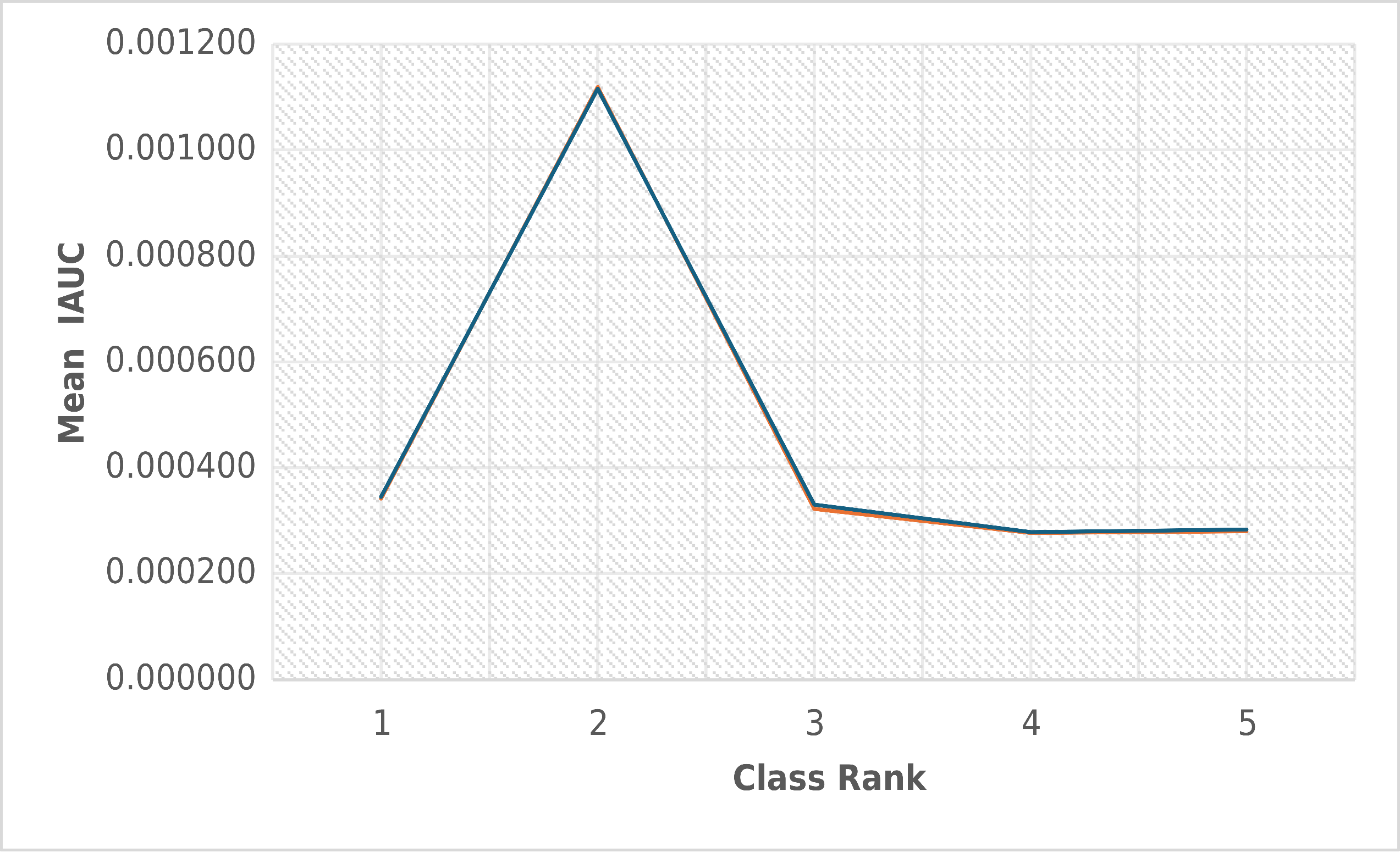}
    \caption{IAUC}
    \end{subfigure}
    \hfill 
    \begin{subfigure}[b]{0.45\linewidth}
    \includegraphics[width=\linewidth]{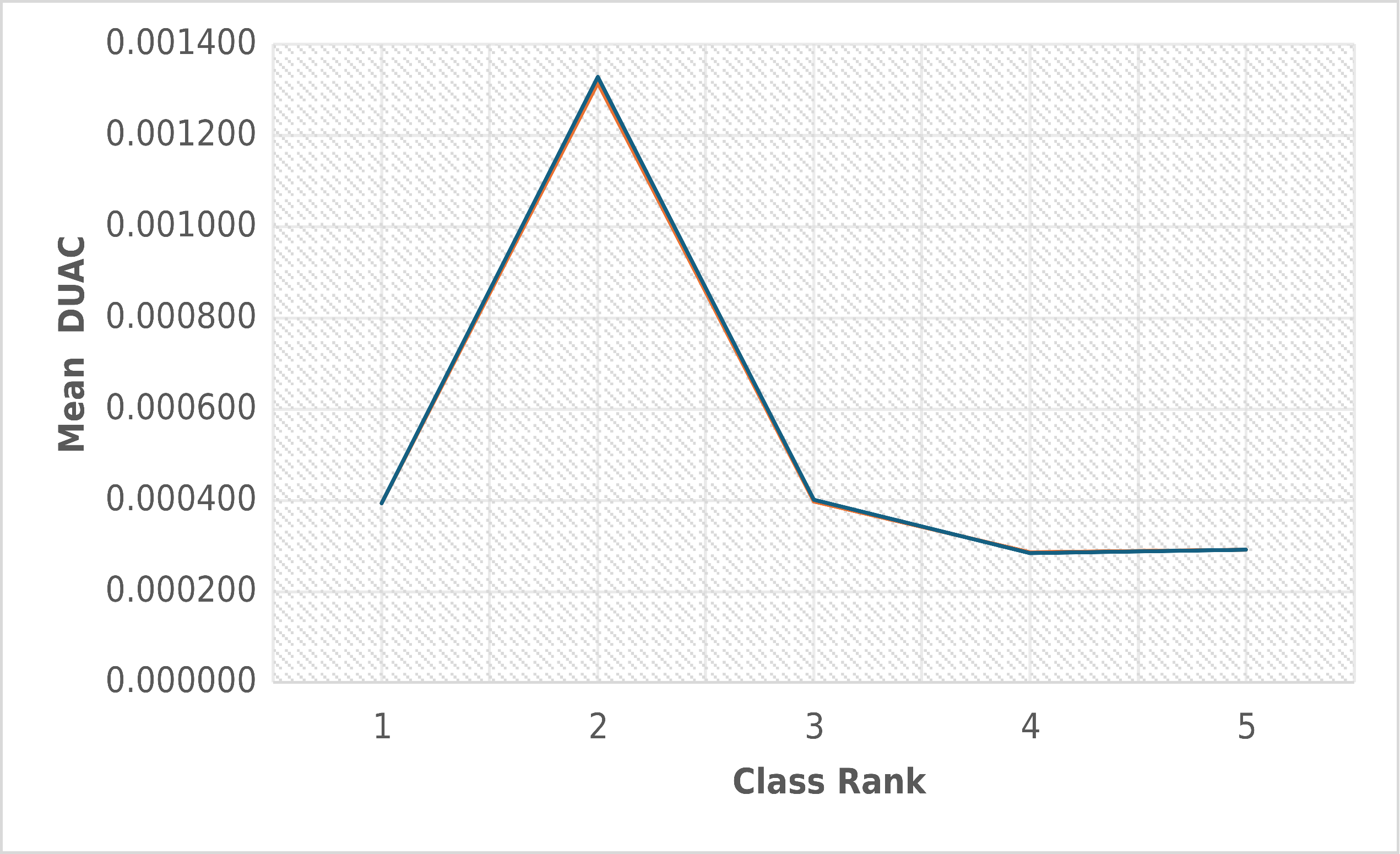}
    \caption{DAUC}
    \end{subfigure}
    \begin{subfigure}[b]{0.45\linewidth}
    \includegraphics[width=\linewidth]{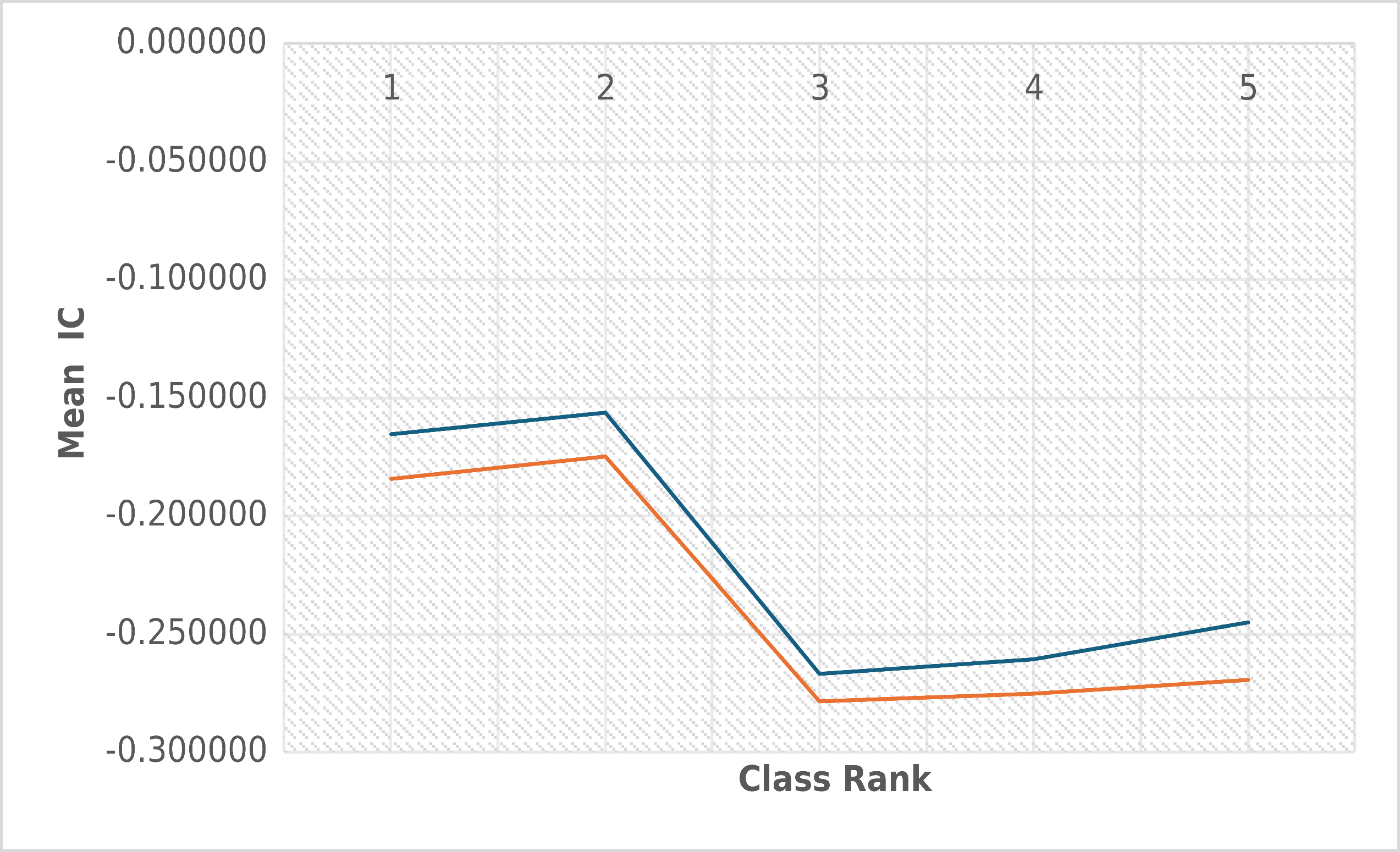}
    \caption{IC}
    \end{subfigure}
    \hfill 
    \begin{subfigure}[b]{0.45\linewidth}
    \includegraphics[width=\linewidth]{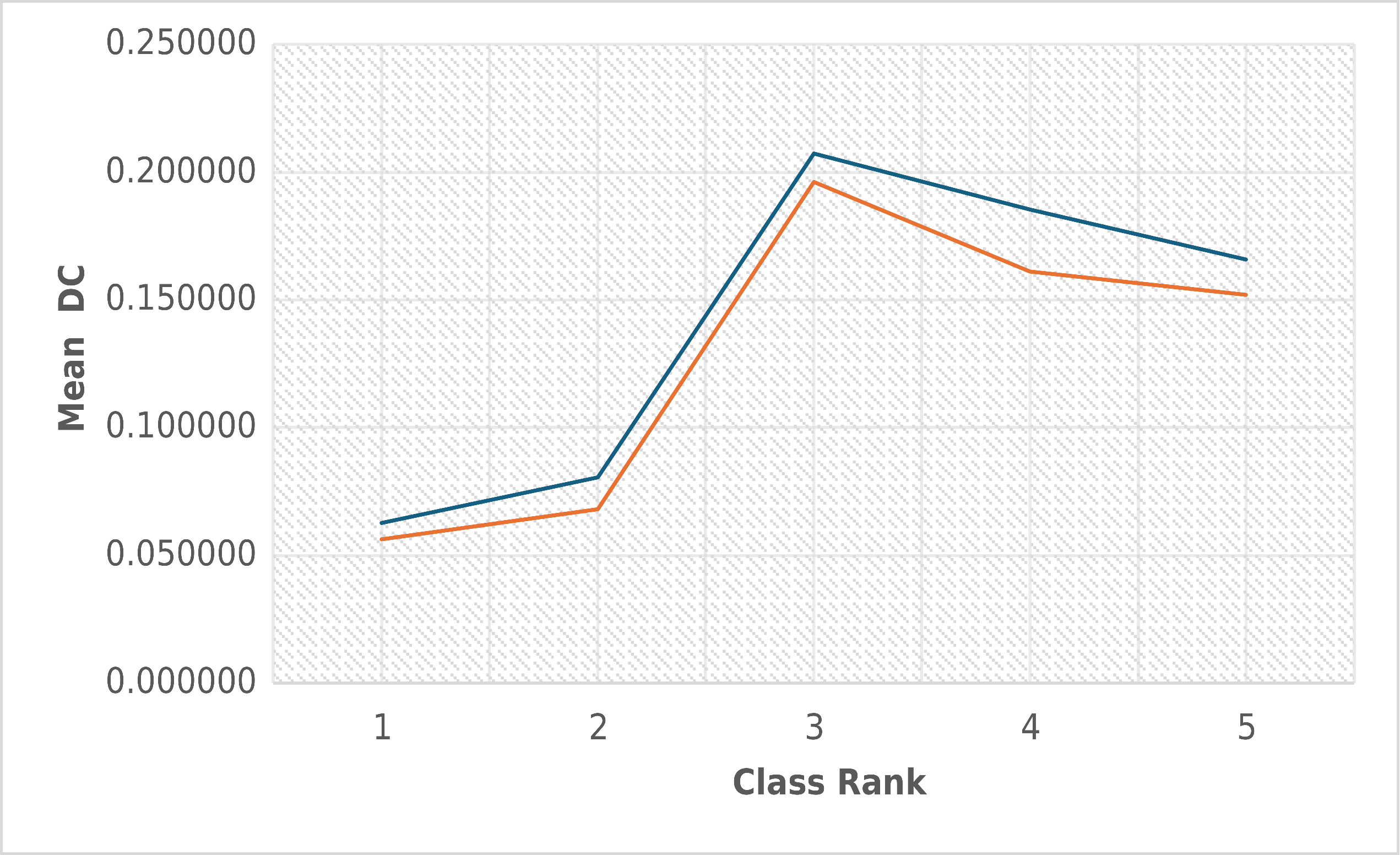}
    \caption{DC}
    \end{subfigure}
    \hfill 
    \begin{subfigure}[b]{0.45\linewidth}
    \includegraphics[width=\linewidth]{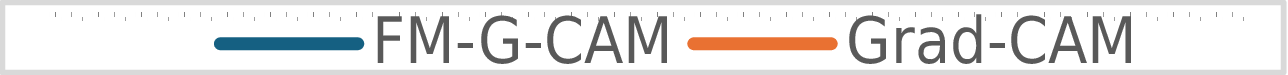}
    \end{subfigure}
    \caption{Results of the quantitative evaluation tests.}
    \label{fig:qe-results}
\end{figure}

In this section, we present the results of the experiments described in the previous section. Table \ref{tbl:qe-numerical-results} summarises the results of the tests carried out, and for most test points, FM-G-CAM achieved higher scores than Grad-CAM. We also graphed this data, as shown in Figure \ref{fig:qe-results}. The figure shows that both techniques perform similarly on the IAUC and DAUC metrics, whereas FM-G-CAM consistently performs better on the IC and DC metrics.

\begin{table}[h]
\centering
\caption{Mean IAUC, IC, DAUC and DC of FM-G-CAM and Grad-CAM calculated for the top-5 predicted classes.}
\label{tbl:qe-numerical-results}
\resizebox{0.6\columnwidth}{!}{%
\begin{tabular}{@{}cccccc@{}}
\toprule
\textbf{CAM Type}                  & \textbf{Class Rank} & \textbf{Mean IAUC} & \textbf{Mean IC}   & \textbf{Mean DAUC} & \textbf{Mean DC}  \\ \midrule
\multirow{5}{*}{\textbf{FM-G-CAM}} & 1                   & \textbf{0.000345}  & \textbf{-0.165391} & 0.000394           & \textbf{0.062597} \\
                                   & 2 & 0.001115          & \textbf{-0.156412} & \textbf{0.001328} & \textbf{0.080689} \\
                                   & 3 & \textbf{0.000330} & \textbf{-0.266839} & \textbf{0.000401} & \textbf{0.207409} \\
                                   & 4 & \textbf{0.000278} & \textbf{-0.260717} & 0.000285          & \textbf{0.185348} \\
                                   & 5 & \textbf{0.000283} & \textbf{-0.244950} & 0.000291          & \textbf{0.165731} \\ \midrule
\multirow{5}{*}{\textbf{Grad-CAM}} & 1 & 0.000342          & -0.184408          & 0.000394          & 0.056323          \\
                                   & 2 & \textbf{0.001119} & -0.174904          & 0.001316          & 0.068046          \\
                                   & 3 & 0.000323          & -0.278517          & 0.000397          & 0.196147          \\
                                   & 4 & 0.000277          & -0.275418          & \textbf{0.000286} & 0.161246          \\
                                   & 5 & 0.000280          & -0.269593          & \textbf{0.000292} & 0.151876          \\ \bottomrule
\end{tabular}%
}
\end{table}


\section{Use cases: A Comparative Practical Guide for FM-G-CAM}
\label{sec:results}

In this section, we compare our method with existing methods for generating saliency maps using a subset of use cases. Notably, the method we propose is fundamentally different from virtually all existing methods because it considers multiple classes in the saliency map. To be as fair as possible, we perform Grad-CAM on all the classes that FM-G-CAM uses to generate the saliency map, as shown in the following sections. We also compare our technique only with Grad-CAM, since any improved or advanced version of Grad-CAM can be used within FM-G-CAM to generate the saliency map. We did not intend to improve the underlying theory of gradient-weighted activations for saliency map generation, but to utilise that idea to provide a more holistic explanation of CNN predictions.
The following sections describe examples of FM-G-CAM use cases in comparison to Grad-CAM.


\subsection{Image Classification}
\begin{figure}[ht]
    \centering
    \includegraphics[width=0.99\textwidth]{general_comparison_1}
    \includegraphics[width=0.99\textwidth]{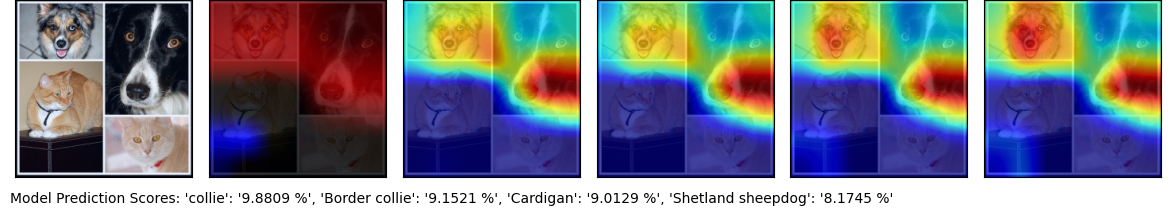}
    \includegraphics[width=0.99\textwidth]{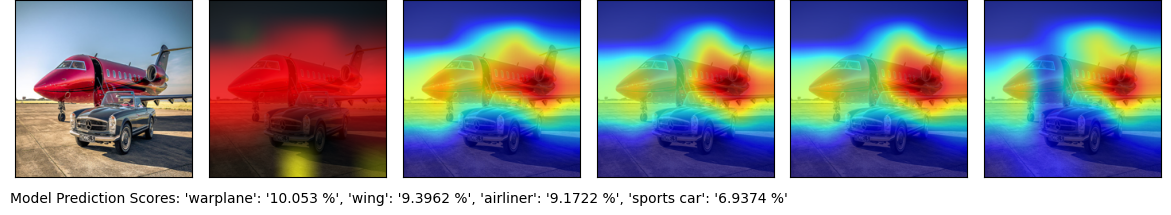}
    \includegraphics[width=0.99\textwidth]{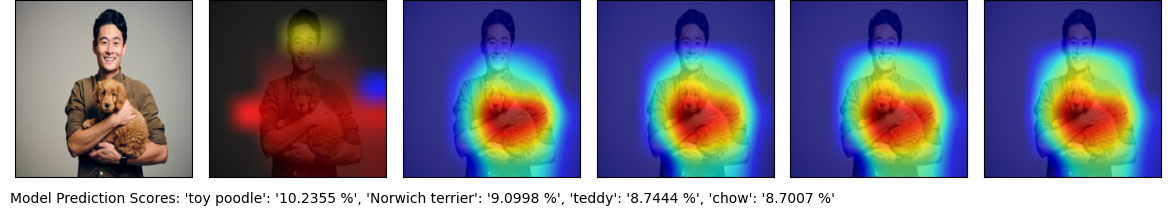}
    \caption{FM-G-CAM for general image classification tasks in comparison with Grad-CAM. Column 2 shows the output for FM-G-CAM, while columns 3 to 6 show the corresponding saliency map outputs from Grad-CAM.}
    \label{fig:general-comparison}
\end{figure}
Figure \ref{fig:general-comparison} shows examples of FM-G-CAM applied to random images. The predictor model used here is ResNet50V2 \cite{he2016identity} with pre-trained ImageNet \cite{deng2009imagenet} weights available in the PyTorch model registry\footnote{\url{https://pytorch.org/vision/stable/models.html}}. The native preprocessing pipeline was used to preprocess the input image.

The prediction in the top row of figure \ref{fig:general-comparison} shows that even when Grad-CAM was generated for the top-4 classes, none of them clearly indicates the presence of the 'tiger cat' class. The proposed approach identifies two main classes that the model has focused on and highlights the areas of these classes in the image. The intensity of the colour represents the importance of the highlighted area for the prediction. The prediction in the bottom row shows an interesting result, since the model's top-4 prediction list does not include a correct label for the human in the image. However, FM-G-CAM shows that there are two separate entities, while all four Grad-CAMs are similar.

\subsection{Medical AI}
\label{subsec:usecase-med-ai}

\begin{figure}[h]
    \centering
    \includegraphics[width=0.99\textwidth]{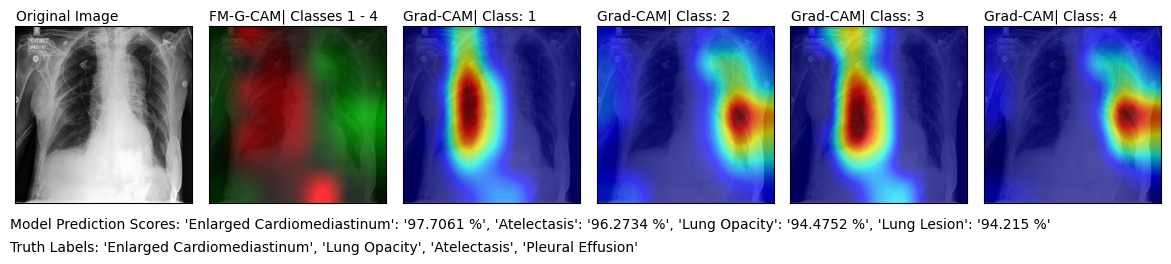}
    \includegraphics[width=0.99\textwidth]{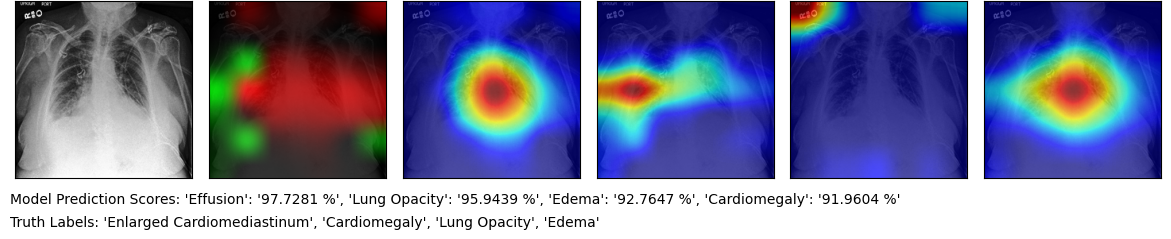}
    \includegraphics[width=0.99\textwidth]{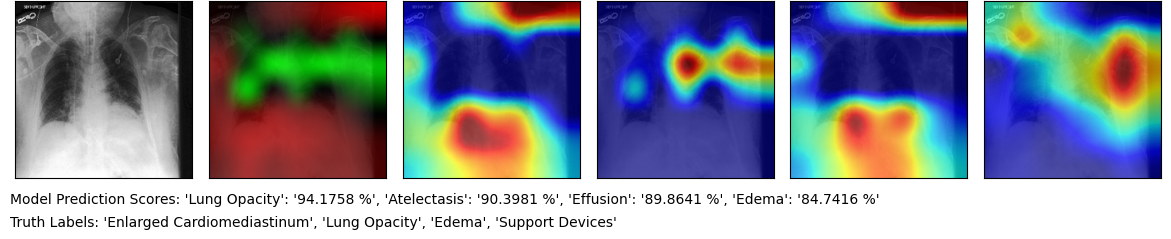}
    \caption{FM-G-CAM for general image classification tasks in comparison with Grad-CAM. Column 2 shows the output for FM-G-CAM, while columns 3 to 6 show the corresponding saliency map outputs from Grad-CAM.}
    \label{fig:xray-comparison}
\end{figure}

Another important area where the proposed approach can be utilised is medical AI. According to the literature, CNNs are a popular approach in the field of medical AI for a variety of diagnostic modalities \cite{sarvamangala2022convolutional, irvin2019chexpert, salehi2020cnn, moon2020computer}. Therefore, an application of FM-G-CAM is explored for the diagnosis of chest X-Ray images. To this end, TorchXRayVision \cite{cohen2022torchxrayvision} weights are used for the DenseNet model \cite{huang2017densely}. Figure \ref{fig:xray-comparison} shows the explained outputs when classifying images selected from the validation set. The samples show that FM-G-CAM produces a holistic, single saliency map that can replace all other class-based Grad-CAM saliency maps. Furthermore, FM-G-CAM visualises the saliency maps of the selected classes comparatively, portraying their relative importance, whereas Grad-CAM considers each class in isolation, completely neglecting the fact that they are all part of the same prediction result.

The clinical significance of this holistic fusion cannot be overstated. In the first sample in Figure \ref{fig:xray-comparison}, the model predicts `Enlarged Cardiomediastinum', `Atelectasis', and `Lung Opacity' with high probabilities ($<$94\%). Clinically, these are not competing, mutually exclusive diagnoses; they frequently co-occur as thoracic pathologies reflecting complex underlying disease states \cite{wang2017chestx, reddy2025cxr}. A standard Grad-CAM approach forces the clinician to view them in isolation, potentially causing them to miss the spatial relationship between the enlarged heart shadow and the adjacent lung opacities. By using FM-G-CAM, the $L_2$-normalised fusion presents a comprehensive pathological footprint. It mirrors the diagnostic process of a human radiologist, who does not assess an X-ray for one disease at a time but rather evaluates the entire radiograph for all contiguous abnormalities simultaneously, thereby aligning the XAI system more closely with user needs and clinical coherence \cite{e2024evaluating}.

\section{Discussion}
\label{sec:discussion}
Using CNN activations to explain predictions was first introduced by \cite{zhou2016learning}, building on earlier work investigating the capabilities of CNNs in object detection \cite{bolei2015object}. These works formed the basis for Grad-CAM \cite{selvaraju2017grad}. However, these studies consider only a single target class and, by doing so, disregard a considerable portion of information regarding the CNN classification process. In this study, we propose that activations should be computed for the top-$k$ classes, weighted by their gradients, normalised for fair representation, and finally fused. Hence, while Grad-CAM and FM-G-CAM are based on the aforementioned studies, they are characteristically and mathematically different.

Figure \ref{fig:general-comparison}, previously discussed, shows that the single-class Grad-CAM approach is not always useful in situations where a bigger-picture approach is needed. Instead, it only shows why a specified class has its related prediction score. In most cases, this class is selected as the top predicted class. The difference between the accuracy of top-1 and the accuracy of top-5 can be substantial \cite{he2016identity, huang2017densely}. Situations such as those arising from the interpretation of chest X-rays show that, in some cases, the top-$k$ predictions where $k>1$ hold significantly useful information when interpreting the predictions.

\section{Conclusion and Future Work}
\label{sec:conclusion}
We highlighted the importance of using holistic visual explanations for CNN predictions in the previous section. It is observed that most saliency map generation methods, such as Grad-CAM, focus solely on a single target class, which can neglect important information related to the reasoning behind a CNN's prediction. In the related literature, we identified that substantial differences between top-1 and top-5 predictions can render saliency maps less relevant when focusing only on the most probable model output. 

To address these gaps, we introduced FM-G-CAM, an unbiased methodology that generates saliency maps capable of providing a comprehensive interpretation of the CNN model's decision-making process. FM-G-CAM can be configured to consider multiple top classes in saliency map generation, providing a more holistic visual explanation for CNN predictions. FM-G-CAM is presented with detailed mathematical and algorithmic explanations, along with a practical implementation guide. Furthermore, to highlight FM-G-CAM's capabilities, we compared its effectiveness with Grad-CAM quantitatively via a customised test plan and qualitatively via real-world use cases. All the code for the approach is released under an open-source licence\footnote{https://github.com/SuienS/cam-evaluation}. 

For future work, the proposed approach uses gradient-weighted activations, although there are other approaches such as Grad-CAM++ \cite{chattopadhay2018grad}. To achieve this, future work could use other methods to provide a further comparison to the proposed FM-G-CAM approach. One limitation of our research is the approach to increasing the visibility of lower activations, which can promote unwanted noise in some cases. Therefore, future work could explore methods to eliminate unwanted noise while retaining the visibility of lower-weighted neuronal activations.

Further analysis could also be performed, such as for the use cases of document analysis \cite{Kosaraju2019}, climate predictions \cite{chattopadhyay2020predicting}, and segmentation tasks \cite{kamnitsas2017efficient, dolz2018hyperdense}, to name a few. The possibility of using FM-G-CAM in object detection could also be investigated.

It is also necessary to consider the computational overhead introduced by FM-G-CAM. Standard Grad-CAM requires a single backward pass to compute gradients for the target class, operating at $O(1)$ with respect to the number of classes. In contrast, FM-G-CAM requires $K$ backward passes to extract gradients for the top-$K$ predictions, resulting in a computational complexity of $O(K)$. While this linear scaling increases inference time for the XAI generation pipeline, the overhead remains minimal in practical deployment scenarios, especially given that $K$ is typically constrained to values $\le 5$. The trade-off, a marginal increase in processing time in milliseconds for a significantly more comprehensive and reliable visual explanation, is highly favourable, particularly in high-stakes domains such as healthcare and autonomous navigation.

To conclude, in this study, we propose a form of explainable artificial intelligence and contribute to the ongoing discussion surrounding more holistic approaches to explainability in image classification problems.

\bibliographystyle{unsrt} 
\bibliography{cas-refs}

@article{arrieta2020explainable,
  title={Explainable Artificial Intelligence (XAI): Concepts, taxonomies, opportunities and challenges toward responsible AI},
  author={Arrieta, Alejandro Barredo and D{\'\i}az-Rodr{\'\i}guez, Natalia and Del Ser, Javier and Bennetot, Adrien and Tabik, Siham and Barbado, Alberto and Garc{\'\i}a, Salvador and Gil-L{\'o}pez, Sergio and Molina, Daniel and Benjamins, Richard and others},
  journal={Information fusion},
  volume={58},
  pages={82--115},
  year={2020},
  publisher={Elsevier}
}

@misc{EUAIAct2024,
  author = {{European Parliament} and {Council of the European Union}},
  title = {{Regulation (EU) 2024/1689} of the {European Parliament} and of the {Council} of 13 {June} 2024 laying down harmonised rules on artificial intelligence and amending {Regulations} {(EC)} {No} 300/2008, {(EU)} {No} 167/2013, {(EU)} {No} 168/2013, {(EU)} 2018/858, {(EU)} 2018/1139 and {(EU)} 2019/2144 and {Directive} {(EU)} 2020/1828 ({Artificial Intelligence Act})},
  journal = {Official Journal of the European Union},
  year = {2024},
  url = {https://eur-lex.europa.eu/eli/reg/2024/1689/oj},
  note = {ELI: http://europa.eu}
}

@article{amann2020explainability,
  title={Explainability for artificial intelligence in healthcare: a multidisciplinary perspective},
  author={Amann, Julia and Blasimme, Alessandro and Vayena, Effy and Frey, Dietmar and Madai, Vince I and Precise4Q Consortium},
  journal={BMC medical informatics and decision making},
  volume={20},
  number={1},
  pages={310},
  year={2020},
  publisher={Springer}
}

@article{minh2022explainable,
  title={Explainable artificial intelligence: a comprehensive review},
  author={Minh, Dang and Wang, H Xiang and Li, Y Fen and Nguyen, Tan N},
  journal={Artificial Intelligence Review},
  pages={1--66},
  year={2022},
  publisher={Springer}
}

@article{adebayo2018sanity,
  title={Sanity checks for saliency maps},
  author={Adebayo, Julius and Gilmer, Justin and Muelly, Michael and Goodfellow, Ian and Hardt, Moritz and Kim, Been},
  journal={Advances in neural information processing systems},
  volume={31},
  year={2018}
}

@article{loh2022application,
  title={Application of explainable artificial intelligence for healthcare: A systematic review of the last decade (2011--2022)},
  author={Loh, Hui Wen and Ooi, Chui Ping and Seoni, Silvia and Barua, Prabal Datta and Molinari, Filippo and Acharya, U Rajendra},
  journal={Computer Methods and Programs in Biomedicine},
  pages={107161},
  year={2022},
  publisher={Elsevier}
}

@inproceedings{barekatain2025evaluating,
  title={Evaluating the explainability of vision transformers in medical imaging},
  author={Barekatain, Leili and Glocker, Ben},
  booktitle={International Workshop on Interpretability of Machine Intelligence in Medical Image Computing},
  pages={96--105},
  year={2025},
  organization={Springer}
}

@article{nafisah2022tuberculosis,
  title={Tuberculosis detection in chest radiograph using convolutional neural network architecture and explainable artificial intelligence},
  author={Nafisah, Saad I and Muhammad, Ghulam},
  journal={Neural Computing and Applications},
  pages={1--21},
  year={2022},
  publisher={Springer}
}

@inproceedings{sudar2022alzheimer,
  title={Alzheimer's Disease Analysis using Explainable Artificial Intelligence (XAI)},
  author={Sudar, K Muthamil and Nagaraj, P and Nithisaa, S and Aishwarya, R and Aakash, M and Lakshmi, S Ishwarya},
  booktitle={2022 International Conference on Sustainable Computing and Data Communication Systems (ICSCDS)},
  pages={419--423},
  year={2022},
  organization={IEEE}
}

@article{alzubaidi2021review,
  title={Review of deep learning: Concepts, CNN architectures, challenges, applications, future directions},
  author={Alzubaidi, Laith and Zhang, Jinglan and Humaidi, Amjad J and Al-Dujaili, Ayad and Duan, Ye and Al-Shamma, Omran and Santamar{\'\i}a, Jos{\'e} and Fadhel, Mohammed A and Al-Amidie, Muthana and Farhan, Laith},
  journal={Journal of big Data},
  volume={8},
  pages={1--74},
  year={2021},
  publisher={Springer}
}

@inproceedings{gupta2022deep,
  title={Deep learning (CNN) and transfer learning: a review},
  author={Gupta, Jaya and Pathak, Sunil and Kumar, Gireesh},
  booktitle={Journal of Physics: Conference Series},
  volume={2273},
  number={1},
  pages={012029},
  year={2022},
  organization={IOP Publishing}
}

@inproceedings{guo2017calibration,
  title={On calibration of modern neural networks},
  author={Guo, Chuan and Pleiss, Geoff and Sun, Yu and Weinberger, Kilian Q},
  booktitle={International conference on machine learning},
  pages={1321--1330},
  year={2017},
  organization={PMLR}
}

@article{han2022survey,
  title={A survey on vision transformer},
  author={Han, Kai and Wang, Yunhe and Chen, Hanting and Chen, Xinghao and Guo, Jianyuan and Liu, Zhenhua and Tang, Yehui and Xiao, An and Xu, Chunjing and Xu, Yixing and others},
  journal={IEEE transactions on pattern analysis and machine intelligence},
  volume={45},
  number={1},
  pages={87--110},
  year={2022},
  publisher={IEEE}
}

@article{khan2022transformers,
  title={Transformers in vision: A survey},
  author={Khan, Salman and Naseer, Muzammal and Hayat, Munawar and Zamir, Syed Waqas and Khan, Fahad Shahbaz and Shah, Mubarak},
  journal={ACM computing surveys (CSUR)},
  volume={54},
  number={10s},
  pages={1--41},
  year={2022},
  publisher={ACM New York, NY}
}

@inproceedings{wang2017chestx,
  title={Chestx-ray8: Hospital-scale chest x-ray database and benchmarks on weakly-supervised classification and localization of common thorax diseases},
  author={Wang, Xiaosong and Peng, Yifan and Lu, Le and Lu, Zhiyong and Bagheri, Mohammadhadi and Summers, Ronald M},
  booktitle={Proceedings of the IEEE conference on computer vision and pattern recognition},
  pages={2097--2106},
  year={2017}
}

@article{Guidotti2018surveryblackbox,
author = {Guidotti, Riccardo and Monreale, Anna and Ruggieri, Salvatore and Turini, Franco and Giannotti, Fosca and Pedreschi, Dino},
title = {A Survey of Methods for Explaining Black Box Models},
year = {2018},
issue_date = {September 2019},
publisher = {Association for Computing Machinery},
address = {New York, NY, USA},
volume = {51},
number = {5},
issn = {0360-0300},
url = {https://doi.org/10.1145/3236009},
doi = {10.1145/3236009},
journal = {ACM Comput. Surv.},
month = {aug},
articleno = {93},
numpages = {42}
}

@inproceedings{deng2009imagenet,
  title={Imagenet: A large-scale hierarchical image database},
  author={Deng, Jia and Dong, Wei and Socher, Richard and Li, Li-Jia and Li, Kai and Fei-Fei, Li},
  booktitle={2009 IEEE conference on computer vision and pattern recognition},
  pages={248--255},
  year={2009},
  organization={Ieee}
}

@article{SOBAHI20221066,
title = {Explainable COVID-19 detection using fractal dimension and vision transformer with Grad-CAM on cough sounds},
journal = {Biocybernetics and Biomedical Engineering},
volume = {42},
number = {3},
pages = {1066-1080},
year = {2022},
issn = {0208-5216},
author = {Nebras Sobahi and Orhan Atila and Erkan Deniz and Abdulkadir Sengur and U. Rajendra Acharya}
}

@InProceedings{Chefer_2021_CVPR,
    author    = {Chefer, Hila and Gur, Shir and Wolf, Lior},
    title     = {Transformer Interpretability Beyond Attention Visualization},
    booktitle = {Proceedings of the IEEE/CVF Conference on Computer Vision and Pattern Recognition (CVPR)},
    month     = {June},
    year      = {2021},
    pages     = {782-791}
}

@article{reddy2025cxr,
  title={CXR-MultiTaskNet a unified deep learning framework for joint disease localization and classification in chest radiographs},
  author={Reddy, K Divya and Patil, Anitha},
  journal={Scientific Reports},
  volume={15},
  number={1},
  pages={32022},
  year={2025},
  publisher={Nature Publishing Group UK London}
}

@article{JUNGIEWICZ2023120234,
title = {Vision Transformer in stenosis detection of coronary arteries},
journal = {Expert Systems with Applications},
volume = {228},
pages = {120234},
year = {2023},
issn = {0957-4174},
doi = {https://doi.org/10.1016/j.eswa.2023.120234},
url = {https://www.sciencedirect.com/science/article/pii/S0957417423007364},
author = {Michal Jungiewicz and Piotr Jastrzebski and Piotr Wawryka and Karol Przystalski and Karol Sabatowski and Stanisław Bartus}
}

@inproceedings{NIPS2012_c399862d,
 author = {Krizhevsky, Alex and Sutskever, Ilya and Hinton, Geoffrey E},
 booktitle = {Advances in Neural Information Processing Systems},
 editor = {F. Pereira and C.J. Burges and L. Bottou and K.Q. Weinberger},
 pages = {},
 publisher = {Curran Associates, Inc.},
 title = {ImageNet Classification with Deep Convolutional Neural Networks},
 url = {https://proceedings.neurips.cc/paper_files/paper/2012/file/c399862d3b9d6b76c8436e924a68c45b-Paper.pdf},
 volume = {25},
 year = {2012}
}

@article{WANG2021203696,
title = {Automated 3D ferrograph image analysis for similar particle identification with the knowledge-embedded double-CNN model},
journal = {Wear},
volume = {476},
pages = {203696},
year = {2021},
note = {23rd International Conference on Wear of Materials},
issn = {0043-1648},
doi = {https://doi.org/10.1016/j.wear.2021.203696},
url = {https://www.sciencedirect.com/science/article/pii/S0043164821000855},
author = {Shuo Wang and Tonghai Wu and Kunpeng Wang}
}

@ARTICLE{Rauber2017Visualization,
  author={Rauber, Paulo E. and Fadel, Samuel G. and Falcão, Alexandre X. and Telea, Alexandru C.},
  journal={IEEE Transactions on Visualization and Computer Graphics}, 
  title={Visualizing the Hidden Activity of Artificial Neural Networks}, 
  year={2017},
  volume={23},
  number={1},
  pages={101-110},
  doi={10.1109/TVCG.2016.2598838}}

@article{Mahendran2016,
	affiliation = {University of Oxford},
	author = {Mahendran, Aravindh and Vedaldi, Andrea},
	copyright = {Springer Science+Business Media New York},
	doi = {10.1007/s11263-016-0911-8},
	journal = {International Journal of Computer Vision},
	language = {English},
	note = {Communicated by Cordelia Schmid.},
	number = {3},
	pages = {233-255},
	title = {Visualizing Deep Convolutional Neural Networks Using Natural Pre-images},
	volume = {120},
	year = {2016},
}

@article{e2024evaluating,
  title={Evaluating explainable artificial intelligence (XAI) techniques in chest radiology imaging through a human-centered lens},
  author={E. Ihongbe, Izegbua and Fouad, Shereen and F. Mahmoud, Taha and Rajasekaran, Arvind and Bhatia, Bahadar},
  journal={Plos one},
  volume={19},
  number={10},
  pages={e0308758},
  year={2024},
  publisher={Public Library of Science San Francisco, CA USA}
}

@INPROCEEDINGS{Bau2017,
  author={Bau, David and Zhou, Bolei and Khosla, Aditya and Oliva, Aude and Torralba, Antonio},
  booktitle={2017 IEEE Conference on Computer Vision and Pattern Recognition (CVPR)}, 
  title={Network Dissection: Quantifying Interpretability of Deep Visual Representations}, 
  year={2017},
  volume={},
  number={},
  pages={3319-3327},
  doi={10.1109/CVPR.2017.354}}

@InProceedings{Zeiler2014,
author="Zeiler, Matthew D.
and Fergus, Rob",
editor="Fleet, David
and Pajdla, Tomas
and Schiele, Bernt
and Tuytelaars, Tinne",
title="Visualizing and Understanding Convolutional Networks",
booktitle="Computer Vision -- ECCV 2014",
year="2014",
publisher="Springer International Publishing",
address="Cham",
pages="818--833"
}

@article{zintgraf2017visualizing,
  title={Visualizing Deep Neural Network Decisions: Prediction Difference Analysis},
  author={Zintgraf, Luisa M and Cohen, Taco S and Adel, Tameem and Welling, Max},
  journal={arXiv e-prints},
  pages={arXiv--1702},
  year={2017}
}

@inproceedings{simonyan2014deep,
  title={Deep inside convolutional networks: visualising image classification models and saliency maps},
  author={Simonyan, K and Vedaldi, A and Zisserman, A},
  booktitle={Proceedings of the International Conference on Learning Representations (ICLR)},
  year={2014},
  organization={ICLR}
}

@inproceedings{springenberg2015striving,
  title={Striving for Simplicity: The All Convolutional Net},
  author={Springenberg, J and Dosovitskiy, Alexey and Brox, Thomas and Riedmiller, M},
  booktitle={ICLR (workshop track)},
  year={2015}
}

@article{bach2015pixel,
  title={On pixel-wise explanations for non-linear classifier decisions by layer-wise relevance propagation},
  author={Bach, Sebastian and Binder, Alexander and Montavon, Gregoire and Klauschen, Frederick and Muller, Klaus-Robert and Samek, Wojciech},
  journal={PloS one},
  volume={10},
  number={7},
  pages={e0130140},
  year={2015},
  publisher={Public Library of Science}
}

@inproceedings{sundararajan2017axiomatic,
  title={Axiomatic attribution for deep networks},
  author={Sundararajan, Mukund and Taly, Ankur and Yan, Qiqi},
  booktitle={International conference on machine learning},
  pages={3319--3328},
  year={2017},
  organization={PMLR}
}

@article{montavon2017explaining,
  title={Explaining nonlinear classification decisions with deep taylor decomposition},
  author={Montavon, Gr{\'e}goire and Lapuschkin, Sebastian and Binder, Alexander and Samek, Wojciech and M{\"u}ller, Klaus-Robert},
  journal={Pattern recognition},
  volume={65},
  pages={211--222},
  year={2017},
  publisher={Elsevier}
}

@inproceedings{zhou2016learning,
  title={Learning deep features for discriminative localization},
  author={Zhou, Bolei and Khosla, Aditya and Lapedriza, Agata and Oliva, Aude and Torralba, Antonio},
  booktitle={Proceedings of the IEEE conference on computer vision and pattern recognition},
  pages={2921--2929},
  year={2016}
}

@inproceedings{selvaraju2017grad,
  title={Grad-cam: Visual explanations from deep networks via gradient-based localization},
  author={Selvaraju, Ramprasaath R and Cogswell, Michael and Das, Abhishek and Vedantam, Ramakrishna and Parikh, Devi and Batra, Dhruv},
  booktitle={Proceedings of the IEEE international conference on computer vision},
  pages={618--626},
  year={2017}
}

@inproceedings{chattopadhay2018grad,
  title={Grad-cam++: Generalized gradient-based visual explanations for deep convolutional networks},
  author={Chattopadhay, Aditya and Sarkar, Anirban and Howlader, Prantik and Balasubramanian, Vineeth N},
  booktitle={2018 IEEE winter conference on applications of computer vision (WACV)},
  pages={839--847},
  year={2018},
  organization={IEEE}
}

@article{fu2020axiom,
  title={Axiom-based grad-cam: Towards accurate visualization and explanation of cnns},
  author={Fu, Ruigang and Hu, Qingyong and Dong, Xiaohu and Guo, Yulan and Gao, Yinghui and Li, Biao},
  journal={arXiv preprint arXiv:2008.02312},
  year={2020}
}

@article{draelos2020use,
  title={Use HiResCAM instead of Grad-CAM for faithful explanations of convolutional neural networks},
  author={Draelos, Rachel Lea and Carin, Lawrence},
  journal={arXiv preprint arXiv:2011.08891},
  year={2020}
}

@InProceedings{Hasany_2023_CVPR,
    author    = {Hasany, Syed Nouman and Petitjean, Caroline and M\'eriaudeau, Fabrice},
    title     = {Seg-XRes-CAM: Explaining Spatially Local Regions in Image Segmentation},
    booktitle = {Proceedings of the IEEE/CVF Conference on Computer Vision and Pattern Recognition (CVPR) Workshops},
    month     = {June},
    year      = {2023},
    pages     = {3733-3738}
}

@INPROCEEDINGS{desaiAblation2020,
  author={Desai, Saurabh and Ramaswamy, Harish G.},
  booktitle={2020 IEEE Winter Conference on Applications of Computer Vision (WACV)}, 
  title={Ablation-CAM: Visual Explanations for Deep Convolutional Network via Gradient-free Localization}, 
  year={2020},
  volume={},
  number={},
  pages={972-980},
  doi={10.1109/WACV45572.2020.9093360}}

@inproceedings{wang2020score,
  title={Score-CAM: Score-weighted visual explanations for convolutional neural networks},
  author={Wang, Haofan and Wang, Zifan and Du, Mengnan and Yang, Fan and Zhang, Zijian and Ding, Sirui and Mardziel, Piotr and Hu, Xia},
  booktitle={Proceedings of the IEEE/CVF conference on computer vision and pattern recognition workshops},
  pages={24--25},
  year={2020}
}

@inproceedings{muhammad2020eigen,
  title={Eigen-cam: Class activation map using principal components},
  author={Muhammad, Mohammed Bany and Yeasin, Mohammed},
  booktitle={2020 international joint conference on neural networks (IJCNN)},
  pages={1--7},
  year={2020},
  organization={IEEE}
}

@article{Jiang2021layercam,
  author={Jiang, Peng-Tao and Zhang, Chang-Bin and Hou, Qibin and Cheng, Ming-Ming and Wei, Yunchao},
  journal={IEEE Transactions on Image Processing}, 
  title={LayerCAM: Exploring Hierarchical Class Activation Maps for Localization}, 
  year={2021},
  volume={30},
  number={},
  pages={5875-5888},
  doi={10.1109/TIP.2021.3089943}}

@inproceedings{collins2018deep,
  title={Deep feature factorization for concept discovery},
  author={Collins, Edo and Achanta, Radhakrishna and Susstrunk, Sabine},
  booktitle={Proceedings of the European Conference on Computer Vision (ECCV)},
  pages={336--352},
  year={2018}
}

@article{Guidotti2018camsurvey,
author = {Guidotti, Riccardo and Monreale, Anna and Ruggieri, Salvatore and Turini, Franco and Giannotti, Fosca and Pedreschi, Dino},
title = {A Survey of Methods for Explaining Black Box Models},
year = {2018},
issue_date = {September 2019},
publisher = {Association for Computing Machinery},
address = {New York, NY, USA},
volume = {51},
number = {5},
issn = {0360-0300},
url = {https://doi.org/10.1145/3236009},
doi = {10.1145/3236009},
month = {aug},
articleno = {93},
numpages = {42}
}

@inproceedings{Liu2018jetcol,
author = {Liu, Yang and Heer, Jeffrey},
title = {Somewhere Over the Rainbow: An Empirical Assessment of Quantitative Colormaps},
year = {2018},
isbn = {9781450356206},
publisher = {Association for Computing Machinery},
address = {New York, NY, USA},
url = {https://doi.org/10.1145/3173574.3174172},
doi = {10.1145/3173574.3174172},
booktitle = {Proceedings of the 2018 CHI Conference on Human Factors in Computing Systems},
pages = {1–12},
numpages = {12},
location = {Montreal QC, Canada},
series = {CHI '18}
}

@INPROCEEDINGS{Morbidelli2020,
  author={Morbidelli, Pietro and Carrera, Diego and Rossi, Beatrice and Fragneto, Pasqualina and Boracchi, Giacomo},
  booktitle={ICASSP 2020 - 2020 IEEE International Conference on Acoustics, Speech and Signal Processing (ICASSP)}, 
  title={Augmented Grad-CAM: Heat-Maps Super Resolution Through Augmentation}, 
  year={2020},
  volume={},
  number={},
  pages={4067-4071},
  doi={10.1109/ICASSP40776.2020.9054416}}

@inproceedings{he2016identity,
  title={Identity mappings in deep residual networks},
  author={He, Kaiming and Zhang, Xiangyu and Ren, Shaoqing and Sun, Jian},
  booktitle={Computer Vision--ECCV 2016: 14th European Conference, Amsterdam, The Netherlands, October 11--14, 2016, Proceedings, Part IV 14},
  pages={630--645},
  year={2016},
  organization={Springer}
}

@article{sarvamangala2022convolutional,
  title={Convolutional neural networks in medical image understanding: a survey},
  author={Sarvamangala, DR and Kulkarni, Raghavendra V},
  journal={Evolutionary intelligence},
  volume={15},
  number={1},
  pages={1--22},
  year={2022},
  publisher={Springer}
}

@inproceedings{irvin2019chexpert,
  title={Chexpert: A large chest radiograph dataset with uncertainty labels and expert comparison},
  author={Irvin, Jeremy and Rajpurkar, Pranav and Ko, Michael and Yu, Yifan and Ciurea-Ilcus, Silviana and Chute, Chris and Marklund, Henrik and Haghgoo, Behzad and Ball, Robyn and Shpanskaya, Katie and others},
  booktitle={Proceedings of the AAAI conference on artificial intelligence},
  volume={33},
  number={01},
  pages={590--597},
  year={2019}
}

@inproceedings{salehi2020cnn,
  title={A CNN model: earlier diagnosis and classification of Alzheimer disease using MRI},
  author={Salehi, Ahmad Waleed and Baglat, Preety and Sharma, Brij Bhushan and Gupta, Gaurav and Upadhya, Ankita},
  booktitle={2020 International Conference on Smart Electronics and Communication (ICOSEC)},
  pages={156--161},
  year={2020},
  organization={IEEE}
}

@article{moon2020computer,
  title={Computer-aided diagnosis of breast ultrasound images using ensemble learning from convolutional neural networks},
  author={Moon, Woo Kyung and Lee, Yan-Wei and Ke, Hao-Hsiang and Lee, Su Hyun and Huang, Chiun-Sheng and Chang, Ruey-Feng},
  journal={Computer methods and programs in biomedicine},
  volume={190},
  pages={105361},
  year={2020},
  publisher={Elsevier}
}

@inproceedings{cohen2022torchxrayvision,
  title={TorchXRayVision: A library of chest X-ray datasets and models},
  author={Cohen, Joseph Paul and Viviano, Joseph D and Bertin, Paul and Morrison, Paul and Torabian, Parsa and Guarrera, Matteo and Lungren, Matthew P and Chaudhari, Akshay and Brooks, Rupert and Hashir, Mohammad and others},
  booktitle={International Conference on Medical Imaging with Deep Learning},
  pages={231--249},
  year={2022},
  organization={PMLR}
}

@inproceedings{huang2017densely,
  title={Densely connected convolutional networks},
  author={Huang, Gao and Liu, Zhuang and Van Der Maaten, Laurens and Weinberger, Kilian Q},
  booktitle={Proceedings of the IEEE conference on computer vision and pattern recognition},
  pages={4700--4708},
  year={2017}
}

@article{bolei2015object,
  title={Object detectors emerge in Deep Scene CNNs},
  author={Bolei, Zhou and Khosla, Aditya and Lapedriza, Agata and Oliva, Aude and Torralba, Antonio},
  year={2015}
}

@INPROCEEDINGS{Kosaraju2019,
  author={Kosaraju, Sai Chandra and Masum, Mohammed and Tsaku, Nelson Zange and Patel, Pritesh and Bayramoglu, Tanju and Modgil, Girish and Kang, Mingon},
  booktitle={2019 International Conference on Document Analysis and Recognition (ICDAR)}, 
  title={DoT-Net: Document Layout Classification Using Texture-Based CNN}, 
  year={2019},
  volume={},
  number={},
  pages={1029-1034},
  doi={10.1109/ICDAR.2019.00168}}

@article{chattopadhyay2020predicting,
  title={Predicting clustered weather patterns: A test case for applications of convolutional neural networks to spatio-temporal climate data},
  author={Chattopadhyay, Ashesh and Hassanzadeh, Pedram and Pasha, Saba},
  journal={Scientific reports},
  volume={10},
  number={1},
  pages={1317},
  year={2020},
  publisher={Nature Publishing Group UK London}
}

@article{kamnitsas2017efficient,
  title={Efficient multi-scale 3D CNN with fully connected CRF for accurate brain lesion segmentation},
  author={Kamnitsas, Konstantinos and Ledig, Christian and Newcombe, Virginia FJ and Simpson, Joanna P and Kane, Andrew D and Menon, David K and Rueckert, Daniel and Glocker, Ben},
  journal={Medical image analysis},
  volume={36},
  pages={61--78},
  year={2017},
  publisher={Elsevier}
}

@article{dolz2018hyperdense,
  title={HyperDense-Net: a hyper-densely connected CNN for multi-modal image segmentation},
  author={Dolz, Jose and Gopinath, Karthik and Yuan, Jing and Lombaert, Herve and Desrosiers, Christian and Ayed, Ismail Ben},
  journal={IEEE transactions on medical imaging},
  volume={38},
  number={5},
  pages={1116--1126},
  year={2018},
  publisher={IEEE}
}

@inproceedings{lin2014microsoft,
  title={Microsoft coco: Common objects in context},
  author={Lin, Tsung-Yi and Maire, Michael and Belongie, Serge and Hays, James and Perona, Pietro and Ramanan, Deva and Doll{\'a}r, Piotr and Zitnick, C Lawrence},
  booktitle={Computer Vision--ECCV 2014: 13th European Conference, Zurich, Switzerland, September 6-12, 2014, Proceedings, Part V 13},
  pages={740--755},
  year={2014},
  organization={Springer}
}

@article{petsiuk2018rise,
  title={Rise: Randomized input sampling for explanation of black-box models},
  author={Petsiuk, Vitali and Das, Abir and Saenko, Kate},
  journal={arXiv preprint arXiv:1806.07421},
  year={2018}
}

@inproceedings{petsiuk2021black,
  title={Black-box explanation of object detectors via saliency maps},
  author={Petsiuk, Vitali and Jain, Rajiv and Manjunatha, Varun and Morariu, Vlad I and Mehra, Ashutosh and Ordonez, Vicente and Saenko, Kate},
  booktitle={Proceedings of the IEEE/CVF Conference on Computer Vision and Pattern Recognition},
  pages={11443--11452},
  year={2021}
}

@inproceedings{gomez2022metrics,
  title={Metrics for saliency map evaluation of deep learning explanation methods},
  author={Gomez, Tristan and Fr{\'e}our, Thomas and Mouch{\`e}re, Harold},
  booktitle={International Conference on Pattern Recognition and Artificial Intelligence},
  pages={84--95},
  year={2022},
  organization={Springer}
}

\end{document}